\documentclass{article}

\PassOptionsToPackage{numbers, compress}{natbib}
\usepackage[final]{nips_2017}
\usepackage{subfiles}

\usepackage[utf8]{inputenc} 
\usepackage[T1]{fontenc}    
\usepackage{hyperref}       
\usepackage{url}            
\usepackage{booktabs}       
\usepackage{amsfonts}       
\usepackage{nicefrac}       
\usepackage{microtype}      
\usepackage{amsmath}
\usepackage{graphicx} 

\usepackage{algorithm}  
\usepackage{algpseudocode}  
\usepackage{amsmath}  

\hypersetup{colorlinks,linkcolor={blue},citecolor={blue},urlcolor={blue}}  
\newcommand{\ourmodel}{EditGAN \citep{ling2021editgan}}
\title{Rethinking the editing of generative adversarial networks: a method to estimate editing vectors based on dimension reduction}

\author{
   Yuhan Cao\\
  \texttt{caoyh1@shanghaitech.edu.cn}
   \And
  Haoran Jiang\\
  \texttt{jianghr1@shanghaitech.edu.cn} \\
   \And
  Zhenghong Yu\\
  \texttt{yuzhh1@shanghaitech.edu.cn}
   \And
   Qi Li\\
  \texttt{liqi2@shanghaitech.edu.cn}
  \And
   Xuyang Li\\
  \texttt{lixy9@shanghaitech.edu.cn}
}

\begin{document}

\maketitle

\begin{abstract}
While Generative Adversarial Networks (GANs) have recently found applications in image editing, most previous GAN-based image editing methods require large-scale datasets with semantic segmentation annotations for training, only provide high level control, or merely interpolate between different images. Previous researchers have proposed  \ourmodel for high-quality, high-precision semantic image editing with limited semantic annotations by finding `editing vectors'. However, it is noticed that there are many features that are not highly associated with semantics, and \ourmodel may fail on them. Based on the orthogonality of latent space observed by \ourmodel, we propose a method to estimate editing vectors that do not rely on semantic segmentation nor differentiable feature estimation network. Our method assumes that there is a correlation between the intensity distribution of features and the distribution of hidden vectors, and estimates the relationship between the above distributions by sampling the feature intensity of the image corresponding to several hidden vectors. We modified Linear Discriminant Analysis (LDA) to deal with both binary feature editing and continuous feature editing. We then found that this method has a good effect in processing features such as clothing type and texture, skin color and hair.

\end{abstract}

\section{Introduction}
\label{gen_inst}
\paragraph{Image Editing and Manipulation.} 

GAN-based image editing methods can be broadly sorted into a number of categories. (i) One line
of work relies on the careful dissection of the GAN’s latent space, aiming to find interpretable and
disentangled latent variables, which can be leveraged for image editing, in a fully unsupervised
manner \citep{012shen2020interpreting,013shen2020interfacegan,014alharbi2020disentangled,024bau2018gan,025bau2020semantic,026plumerault2020controlling,027harkonen2020ganspace,047goetschalckx2019ganalyze,048jahanian2019steerability,049voynov2020unsupervised,050wang2021geometry,051shen2021closed}. Although powerful, these approaches usually do not result in any
high-precision editing capabilities. The editing vectors we are learning in \ourmodel would be too
hard to find independently without segmentation-based guidance. (ii) Other works utilize GANs
that condition on class or pixel-wise semantic segmentation labels to control synthesis and achieve
editing \citep{009choi2018stargan,010lee2020maskgan,011wu2020cascade,019zhu2020sean,022chen2020deepfacedrawing,046park2019semantic,052wang2018high}. Hence, these works usually rely on large annotated datasets, which are
often not available, and even if available, the possible editing operations are tied to whatever labels are
available. This stands in stark contrast to \ourmodel, which can be trained in a semi-supervised fashion
with very little labeled data and where an arbitrary number of high-precision edits can be learnt. (iii)
Furthermore, auxiliary attribute classifiers have been used for image manipulation \citep{015hou2022guidedstyle,023he2019attgan}, thereby
still relying on annotated data and usually only providing high-level control. (iv) Image editing is
often explored in the context of “interpolating” between a target and different reference image in
sophisticated ways, for example by replacing certain features in a given image with features from a
reference images \citep{018collins2020editing,019zhu2020sean,020lewis2021vogue,021kim2021exploiting}. From the general image editing perspective, the requirement of reference
images limits the broad applicability of these techniques and prevents the user from performing
specific, detailed edits for which potentially no reference images are available. (v) Recently, different
works proposed to directly operate in the parameter space of the GAN instead of the latent space to
realize different edits \citep{016cherepkov2021navigating,025bau2020semantic,028bau2020rewriting}. For example, \cite{025bau2020semantic,028bau2020rewriting} essentially specialize the generator network
for certain images at test time to aid image embedding or “rewrite” the network to achieve desired
semantic changes in output. The drawback is that such specializations prevent the model from being
used in real-time on different images and with different edits. \citep{016cherepkov2021navigating} proposed an approach that more
directly analyses the parameter space of a GAN and treats it as a latent space in which to apply
edits. However, the method still merely discovers edits in the network’s parameter space, rather than
actively defining them like we do. It remains unclear whether their method can combine multiple
such edits, as we can, considering that they change the GAN parameters themselves. (vi) Finally,
another line of research targets primarily very high-level image and photo stylization and global
appearance modifications \citep{037gatys2016image,041park2020swapping,046park2019semantic,052wang2018high,053luan2017deep,054liu2017unsupervised,055li2018closed,056kazemi2019style,057yoo2019photorealistic}.

Generally, most works only do relatively high-level and not the detailed, high-precision editing,
which our model targets. Hence, we consider our model as complementary to this body of work.

\paragraph{GANs and Latent Space Image Embedding.} \ourmodel builds on top of DatasetGAN \cite{001zhang2021datasetgan} and
SemanticGAN \citep{002li2021semantic}, which proposed to jointly model images and their semantic segmentations using
shared latent codes. However, these works leveraged this model design only for semi-supervised
learning, not for editing. \ourmodel also relies on an encoder, together with optimization, to embed
new images to be edited into the GAN’s latent space. This task in itself has been studied extensively in
different contexts before, and we are building on these works. Previous papers studied encoder-based
methods \citep{058perarnau2016invertible,059donahue2016adversarial,060brock2016neural,061dumoulin2016adversarially,062richardson2021encoding}, used primarily optimization-based techniques \citep{026plumerault2020controlling,063zhu2016generative,064yeh2017semantic,065lipton2017precise,066abdal2019image2stylegan,067huh2020transforming,068creswell2018inverting,069raj2019gan}, and developed hybrid
approaches \cite{024bau2018gan,025bau2020semantic,063zhu2016generative,070bau2019seeing,071zhu2020domain}.

Finally, a concurrent paper \citep{072xu2021linear} shares similarities with DatasetGAN \cite{001zhang2021datasetgan}, on which our method builds,
and explores an editing approach related to \ourmodel as one of its applications. However, our
editing approach is methodologically different and leverages editing vectors, and also demonstrates
significantly more diverse and stronger experimental results. Furthermore, \citep{073bau2021paint} shares some high level ideas with \ourmodel; however, it leverages the CLIP \citep{074radford2021learning} model and targets text-driven editing.

\section{Related work}
\label{headings}
\subsection{GAN and latent space image embedding}
DatasetGAN \citep{001zhang2021datasetgan} and SemanticGAN \citep{002li2021semantic} proposed to jointly model images and their semantic segmentations using shared latent codes. For the generative adversarial network model, 
$$\textnormal{Generator: } \mathcal{W^{+}}\rightarrow \mathcal{X}$$
$$\textnormal{Discriminator: } \mathcal{X}\rightarrow \mathrm{score}[0,1]$$
The generator generates pictures of image space $\mathcal{X}$ from latent space $\mathcal{W^{+}}$, while the discriminator classified the generated pictures to different label. However, these works leveraged this model design only for semi-supervised learning, not for editing.
\subsection{EditGAN}

\begin{figure}[htbp] 
	\begin{center} 
		\includegraphics[width=0.9\textwidth]{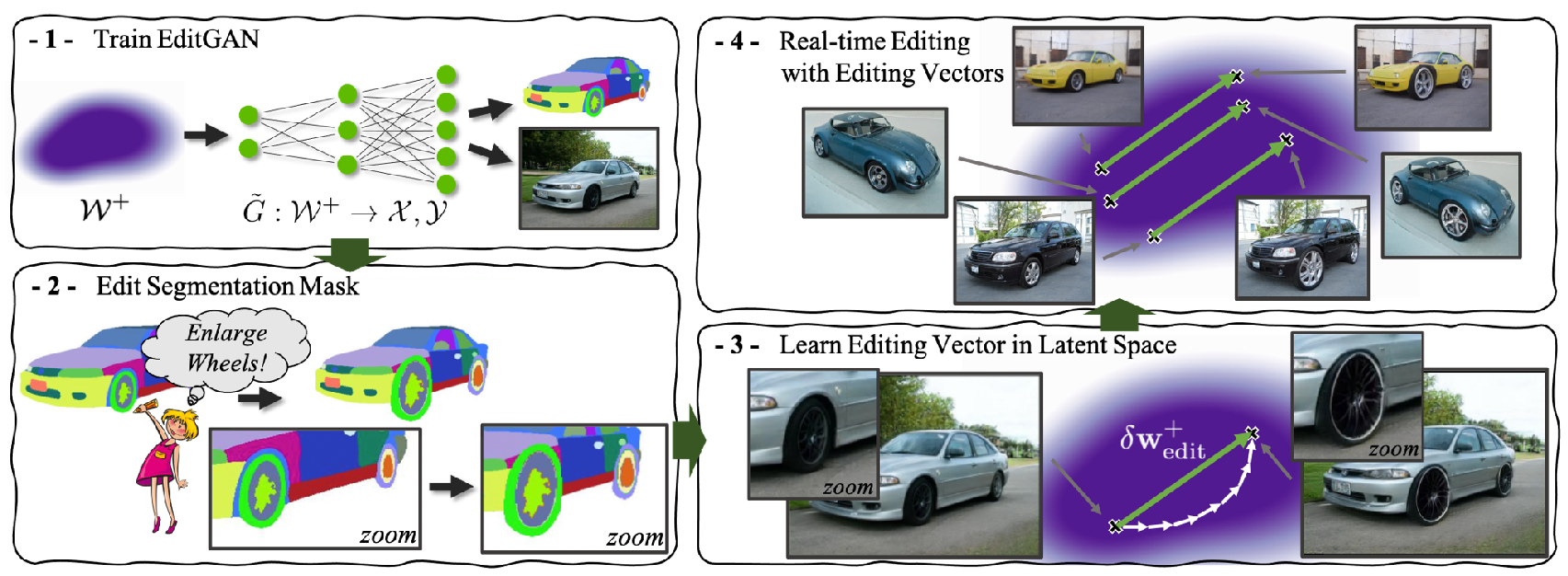}
	\end{center} 
	\caption{Pipeline of EditGAN}
	\label{fig:continuousresult} 
	\vspace{-10pt}
\end{figure}
Our model lies in leveraging the joint distribution of images and semantic segmentation for high-precision image editing. EditGAN try to embed new image into latent space, generate the corresponding segmentation output. The new model is,
$$\textnormal{Generator: } \mathcal{W^{+}}\rightarrow \mathcal{X}, \mathcal{Y}$$
$$\textnormal{Discriminator: } \mathcal{X}, \mathcal{Y}\rightarrow \mathrm{score}[0,1]$$
while $\mathcal{Y}$ represents semantic space. \ourmodel denote the edited segmentation mask by $\mathbf{y}_{\textnormal{edited}}$. Starting from the embedding $\mathbf{w}^{+}$ of the unedited image $\mathbf{x}$ and segmentation $\mathbf{y}$, \ourmodel then perform optimization within $\mathcal{W}^{+}$ to find a new $\mathbf{w}^{+}_{\textnormal{edited}}=\mathbf{w}^{+}+\delta\mathbf{w}^{+}_{\textnormal{edit}}$ consistent with the new segmentation $\mathbf{y}_{\textnormal{edited}}$, while allowing the RGB output $\mathbf{x}$ to change within the editing region.

\ourmodel formally seeks an $editing\ vector$ $\delta\mathbf{w}_{edit}^{+}\in\mathcal{W}^{+}$ such that $(\mathbf{x}_{edited}, \mathbf{y}_{edited} = \tilde{G}(\mathbf{w}^{+} + \delta\mathbf{w}_{edit}^{+})$, where $\tilde{G}$ denotes the fixed generator that synthesizes both images and segmentation. To find $\delta\mathbf{w}^{+}$, approximating $\delta\mathbf{w}_{edit}^{+}$, \ourmodel use the following losses as minimization targets:
\begin{equation}
    \begin{split}
        \mathcal{L}_{\textnormal{RGB}}(\delta\mathbf{w}^{+})
        &=L_{\textnormal{LPIPS}}(\tilde{G}^{\mathbf{x}}(\mathbf{w}^{+}+\delta\mathbf{w}^{+})\odot(1-r), \mathbf{x}\odot (1-r)) \\
        &+L_{L2}(\tilde{G}^{\mathbf{x}}(\mathbf{w}^{+}+\delta\mathbf{w}^{+})\odot(1-r), \mathbf{x}\odot (1-r))
    \end{split}
\end{equation}

\begin{equation}
    \mathcal{L}_{\textnormal{CE}}(\delta\mathbf{w}^{+})=H(\tilde{G}^{\mathbf{y}}(\mathbf{w}^{+}+\delta\mathbf{w}^{+})\odot r, \mathbf{y}_{\textnormal{edited}}\odot r)
\end{equation}

\begin{equation}
    \mathcal{L}_{\textnormal{ID}}(\delta\textbf{w}^{+}) = \langle R(\tilde{G}^{\textbf{x}}(\textbf{w}^{+}+\delta\textbf{w}^{+}),R(\textbf{x}))\rangle 
\end{equation}

where $H$ denotes the pixel-wise cross-entropy, $L_{\textnormal{LPIPS}}$ loss is base on the Learned Perception Image Patch Similarity (LPIPS) distance \citep{075zhang2018unreasonable}, and $L_{L2}$ is a regular pixel-wise L2 loss. $\mathcal{L}_{\textnormal{RGB}}(\delta\mathbf{w}^{+})$ ensures that the image appearance does not change $outside$ the region of interest, while $\mathcal{L}_{\textnormal{CE}}(\delta\mathbf{w}^{+})$ ensures that the target segmentation $\mathbf{y}_{\textnormal{edit}}$ is enforced $within$ the editing region. When editing human faces, \ourmodel also apply the identity loss $\mathcal{L}_{\textnormal{ID}}(\delta\mathbf{w}^{+})$, with $R$ denoting the pre-trained ArcFace feature extraction network and $\langle \cdot,\cdot\rangle$ cosine-similarity.

The final objective function for optimization then becomes:
\begin{equation}
    \mathcal{L}_{\textnormal{editing}}(\delta\mathbf{w}^{+}) = \lambda_{1}^{\textnormal{editing}}\mathcal{L}_{\textnormal{RGB}}(\delta\mathbf{w}^{+})+\lambda_{2}^{\textnormal{editing}} \mathcal{L}_{\textnormal{CE}}(\delta\mathbf{w}^{+})+\lambda_{3}^{\textnormal{editing}}  \mathcal{L}_{\textnormal{ID}}(\delta\mathbf{w}^{+})
\end{equation}

with hyper-parameters $\lambda_{1,\cdots,3}^{\textnormal{editing}}$. The only "learnable" variable is the editing vector $\delta\mathbf{w}^{+}$; all neural networks are kept fix. After optimizing $\delta\mathbf{w}^{+}$ with the objective function, we can use $\delta\mathbf{w}^{+}\approx \delta\mathbf{w}_{edit}^{+}$. \ourmodel rely on the generator, trained to synthesize realistic images, to modify the RGB values in the editing region in a plausible way consistent with the segmentation edit.

Summarizing, \ourmodel  perform image editing with \ourmodel in three different modes:
\begin{itemize}
  \item \textbf{Real-time Editing with Editing Vectors.} For localized, well-disentangled edits \ourmodel perform editing purely by applying previously learnt editing vectors with varying scales $s_{edit}$ and manipulate images at interactive rates.
  \item \textbf{Vector-based Editing with Self-Supervised Refinement.} For localized edits that are not perfectly disentangled with other parts of the image, we can remove editing artifacts by additional optimization at test time, while initializing the edit using the learnt editing vectors.
  \item \textbf{Optimization-based Editing.} Image-specific and very large edits do not transfer to other images via editing vectors. For such operations, \ourmodel perform optimization from scratch.
\end{itemize} 

\subsection{Shortage of EditGAN}
Although \ourmodel pays attention to the orthogonality of latent space, it only uses this property to propose the method of editing vectors, and does not make use of the properties of space in the process of solving editing vectors.

Besides, as we have introduced above, \ourmodel is an optimization based approach which requires plenty handful annotations and heavy training, especially when training its semantic branch.
\section{Method}
\subsection{Latent space analysis}
InterfaceGAN \citep{013shen2020interfacegan} proposed that hyperplane can be found in latent space to classify whether the image generated by any latent vector has specific features, which inspired us to consider the relationship between the hyperplane found in latent space to classify features and the corresponding editing vector.

Given a well-trained GAN model, the generator can be viewed as a deterministric function $g:\ \mathcal{W}\rightarrow \mathcal{X}$. Here, $\mathcal{W}\in \mathbb{R}$ denotes the $d$-dimensional latent space. $\mathcal{X}$ stands for the image space, where each sample $\textbf{x}$ possesses certain semantic information, like gender and age for face model. Suppose we have a semantic scoring function $f_{s}:\ \mathcal{X}\rightarrow \mathcal{S}$, where $\mathcal{S}\in \mathbb{R}^{m}$ represents the semantic space with $m$ semantics. InterfaceGAN \citep{013shen2020interfacegan} bridge the latent space $\mathcal{W}$ and the semantic space $\mathcal{S}$ with $\textbf{s} = f_{s}(g(\textbf{z}))$, where $\textbf{s}$ and $\textbf{z}$ denote semantic scores and the sampled latent code respectively.  

Given a hyperplane with unit normal vector $\textbf{n}\in\mathbb{R}^{d}$, we define the "distance" from a sample $\textbf{z}$ from lantent space to this hyperplane as
\begin{equation}
    d(\textbf{n}, \textbf{z}) = \textbf{n}^{T}\textbf{z}
\end{equation}
Here, $d(\cdot,\cdot)$ is not a strictly defined distance since it can be negative. When \textbf{z} lies near the boundary and across the hyperplane, both the "distance" and the semantic score very accordingly. Moreover, it is just when the "distance" changes its numerical sign that the semantic attribute reverses. We therefore expect these two items to be linearly dependent with
\begin{equation}
    f(g(\textbf{z})) = \lambda d(\textbf{\textbf{n}, \textbf{z}})
\end{equation}
where $f(\cdot)$ is the scoring function for a particular semantic, and $\lambda >0$ is a scalar to measure how fast the semantic varies along with the change of "distance". Random samples drawn from specific distribution are very likely to locate close enough to a given hyperplane. Therefore, the corresponding semantic can be modeled by the linear subspace that is defined by $\textbf{n}$. 

One of the major differences between our case and InterfaceGAN \citep{013shen2020interfacegan} is the optimization problem. InterfaceGAN \citep{013shen2020interfacegan} uses Support Vector Machines (SVM) \citep{cortes1995support} to classify existence of feature in latent space, which is to maximize interval between two classes. In our case, the objective function is slightly different, which is to find the vector from the center of one class to another.

Another major difference is the number of annotations. InterfaceGAN \citep{013shen2020interfacegan} originally uses $6000$ fine-annotated samples (12 times of the dimension of latent space), while in we uses only $1000$ samples (2 times of the dimension of latent space). Which lead us to dimension reduction methods rather than learning base methods.

\subsection{Dimension reduction in latent space}
The features we deal with fall into two categories: binary features which has only two states (have or not have); continuous features which can be regarded to has strength from 0 to 1. 
\subsubsection{Binary feature editing}
With respect to binary features, We introduced Linear Discrimination Analysis(LDA) \citep{riffenburgh1957linear} to handle binary feature. In the binary classification problem, LDA \citep{riffenburgh1957linear} can robustly estimate the projection vector that maximizes the distance between classes and minimizes the variance within a class after data points are projected.

\subsubsection{Continuous features editing}
Estimating editing vector for continuous features is far more tricky. Here we proposed two approaches that are bipolar method and discretizing method.

Bipolar method is to split the dataset into low, medium and high strength parts, and let the low strength part be class 0 and high strength part be class 1 to obtain estimation of editing vector. This method is quite straightforwards but not accurate nor efficient.

Discretizing method also split the dataset into parts but with more bins. Ordinarily we use a setting with $5$ groups: $[0,0.2)$, $[0.2,0.4)$, $[0.4,0.6)$, $[0.6,0.8)$ and $[0.8,1]$. Rather than performing multi-class LDA \citep{riffenburgh1957linear} to obtain a class by class discriminator, we set the optimization problem to find a single projection vector that maximize between class scatter and minimize within class scatter when projected on it. The normal form of the optimization problem is define as below:

\subsubsection{Modified Linear Discriminant Analysis}
Given a dataset $\mathbf{X}$ with $M$ labels, we can separate the dataset into several groups according to the labels, which is denoted as $\mathbf{X}_i$, where $i \in \{1,...,M\}$. In each group $\mathbf{X}_i$ of size $n_i$, we denote $x_{ij} \in \mathbb{R}^N$ as the data point in this group where $j \in \{1,...,n_i\}$ and $m_{i}$ as the group center, i.e. $m_{i} = \frac{1}{\mathbf{n_i}}\sum_j x_{ij}$. Denote $\Bar{m}$ as the center of the whole dataset $\mathbf{X}$. 

In order to get a good classification result, we need to maximize variance between groups $Var_{bg}$ and minimize variance in groups $Var_{ig}$ after projection. Denote $\mu \in \mathbb{R}^N$ as the projection vector. Hence, we can calculate both two variances. 
\begin{equation}
    \begin{aligned}
        Var_{bg} = \sum_{i=1}^{M}\left[ \mu^T(m_i- \Bar{m})\right]^2 = \mu^T \left(\sum_{i=1}^{M} (m_i-\Bar{m})(m_i-\Bar{m})^T \right)\mu
    \end{aligned}
\end{equation}
\begin{equation}
    Var_{ig} = \sum_{i=1}^{M} \sum_{j=1}^{n_i} \left[\mu^T(x_{ij}-m_i)\right]^2 = \mu^T \left(\sum_{i=1}^{M} \sum_{j=1}^{n_i}(x_{ij}-m_i)(x_{ij}-m_i)^T\right)\mu
\end{equation}
For convenience, we denote $V = \sum_{i=1}^{M} \sum_{j=1}^{n_i}(x_{ij}-m_i)(x_{ij}-m_i)^T$ and $S = \sum_{i=1}^{M} (m_i-\Bar{m})(m_i-\Bar{m})^T$. 

In order to maximize $Var_{bg}$ and minimize $Var_{ig}$ at the same time, we formulate the following optimization problem:
\begin{eqnarray}
    \max_{\mu} \qquad \frac{\mu^T V \mu}{\mu^T S \mu} \label{pre_opt}
\end{eqnarray}
It is trivial to see that \eqref{pre_opt} is equivalent to the following:
\begin{eqnarray}
\begin{aligned}
    \max_{\mu} & \qquad \mu^T V \mu \\
    \mathrm{s.t.} & \qquad \mu^TS\mu = 1.
\end{aligned}
\end{eqnarray}
By applying Lagrange multipliers, we could define the Lagrange function $L$:
\begin{equation}
    L = \mu^T V \mu - \lambda(\mu^TS\mu - 1)
\end{equation}
According to the KKT conditions, we could get the optimal value is the biggest eigenvector $\lambda_1$ of $S^{-1}V$ and the corresponding solution $\mu^*$ is the corresponding eigenvector of $\lambda_1$.

Noting that $S$ may not be invertible, which could block the calculation. We could use some technique to avoid this situation. Denote $\epsilon$ as a small number and $I \in \mathbb{R}^{N\times N}$ as the identity matrix. We add $\epsilon I$ to $S$ to keep the new $S$ invertible.

\section{Experiments}
As GAN based editing methods require a well-trained GAN network as base model, we choose StyleGAN-Human as our base GAN model.
\subsection{EditGAN methods}
\begin{figure}[htbp] 
	\begin{center} 
		\includegraphics[height=25mm]{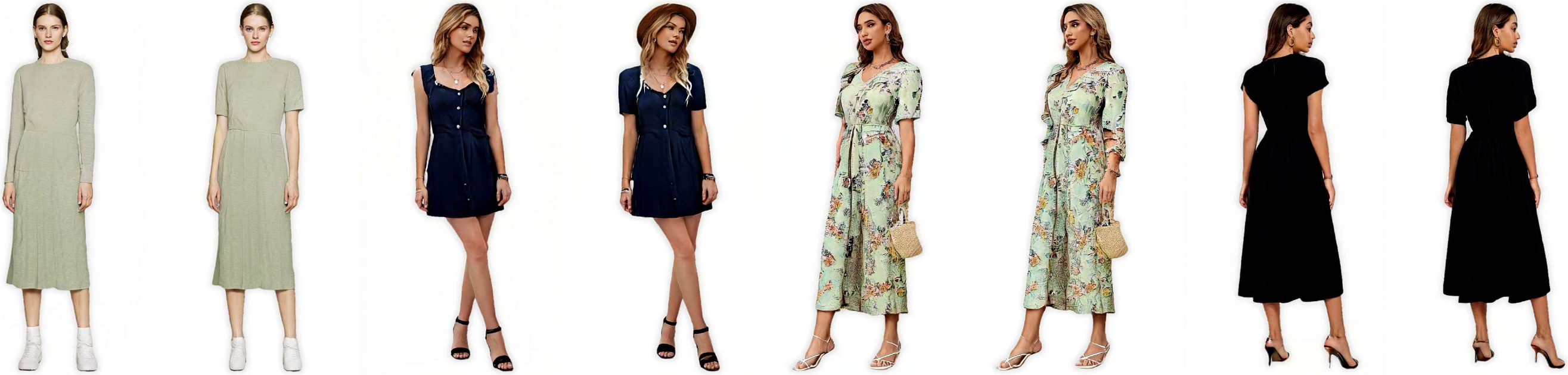}
	\end{center} 
	\caption{Editing results of \ourmodel methods.}
	\label{fig:editganresult} 
	\vspace{-10pt}
\end{figure}

According to the method of \ourmodel, we manually labeled 100 groups of human body segmentation data, trained semantic branches, and then obtained the editing vectors of the upper garment lengths. The editing results by vectors obtained by \ourmodel methods are shown in Fig \ref{fig:editganresult}.

\begin{figure}[htbp] 
	\begin{center} 
		\includegraphics[height=25mm]{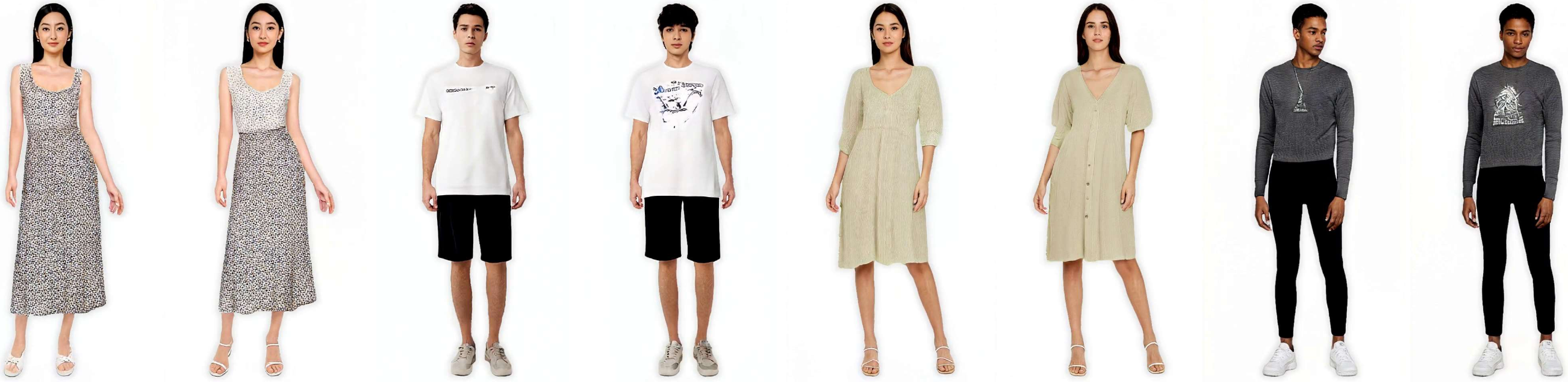}
	\end{center} 
	\caption{Editing results of binary feature}
	\label{fig:binaryresult} 
	\vspace{-10pt}
\end{figure} 

\begin{figure}[htbp] 
	\begin{center} 
		\includegraphics[height=25mm]{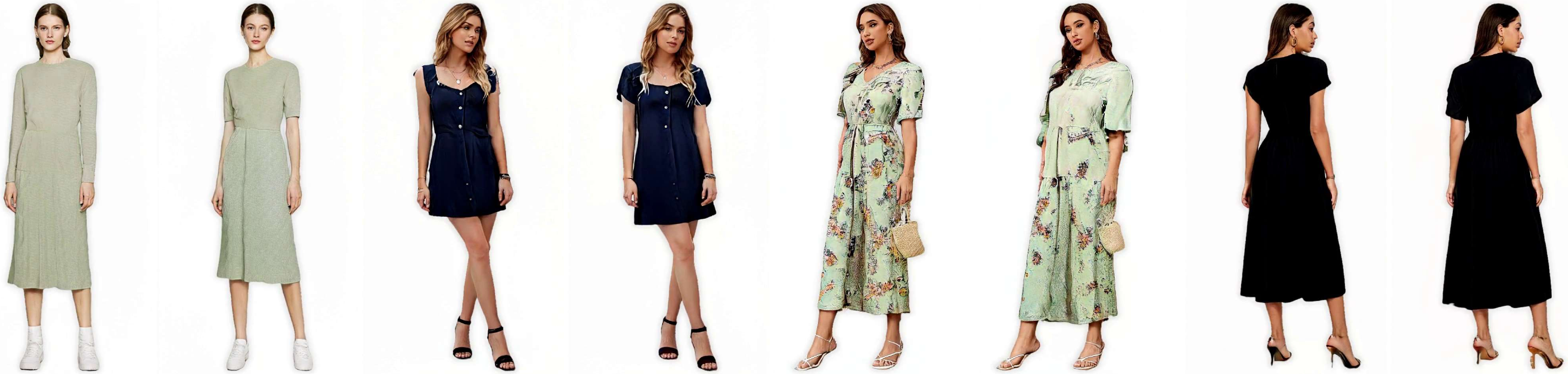}
	\end{center} 
	\caption{Editing results of continuous feature}
	\label{fig:continuousresult} 
	\vspace{-10pt}
\end{figure}
\begin{table}[htbp]
  \caption{Cross validation results between EditGAN and ours}
  \label{table:crossvalidation} 
  \centering
  \begin{tabular}{lll}
    \toprule
    Metrics  & EditGAN\citep{ling2021editgan}& Ours \\
    \midrule
    Correlation & 0.5788  & 0.3911     \\
    L2 distance & \multicolumn{2}{c}{1.1214}        \\
    \bottomrule
  \end{tabular}
\end{table}
\subsection{Proposed methods}
\paragraph{Binary feature editing}
We set whether there is pattern on cloth as the target feature and labeled 500 images. According to the method discussed above, we perform LDA \citep{riffenburgh1957linear} over the sample sets and obtain the editing vector. The editing results are shown in Fig \ref{fig:binaryresult}. 
\paragraph{Continuous feature editing}
We use the length of upper garment as target feature to compare with \ourmodel method and labeled 500 images. Then the discretizing method was performed according to the method discussed above to obtain the editing vector. The editing results are shown in Fig \ref{fig:continuousresult}.
\subsection{Cross validation}
As both methods estimate the editing vector for length of upper garment, it is possible to compare their effectiveness on feature strength estimation and consistency. We compared the correlation between projection length and feature strength and the L2 distance between two editing vector. The results are shown in Table \ref{table:crossvalidation}.

\section{Conclusion}
We have developed an efficient GAN image editing technology based on the vector space editing technology proposed by \ourmodel, and proved that our method can efficiently find the general direction of the editing vector under its limited annotation data. Compared with \ourmodel, our method explicitly has the ability to edit discrete binomial features; Compared with InterfaceGAN \citep{013shen2020interfacegan}, our method proposes an efficient method to find edit vectors on continuous features. At the same time, we noticed that the editing vector we found was not completely decoupled from other features, and some other features also changed in the editing process.
In the future, we plan to explore more explicit editing vector estimation methods on continuous features. We also plan to better solve the decoupling problem of editing vectors in finite annotation.
\section{Contribution percent}

Haoran Jiang: 20\%

Qi Li: 20\%

Xuyang Li: 20\%

Yuhan Cao: 20\%

Zhenghong Yu: 20\%

\bibliographystyle{IEEEtran}
\bibliography{egbib}
\end{document}